\documentclass[letterpaper]{article} 
\usepackage{aaai25}  
\usepackage{times}  
\usepackage{helvet}  
\usepackage{courier}  
\usepackage[hyphens]{url}  
\usepackage{graphicx} 
\usepackage{natbib}  
\usepackage{caption} 
\usepackage{algorithm}
\usepackage{algorithmic}
\usepackage{amsmath}
\usepackage{amssymb}
\usepackage{booktabs}
\usepackage{multirow}
\usepackage{xcolor}
\usepackage{enumitem}

\usepackage{newfloat}
\usepackage{listings}
\DeclareCaptionStyle{ruled}{labelfont=normalfont,labelsep=colon,strut=off} 
\floatstyle{ruled}
\newfloat{listing}{tb}{lst}{}
\floatname{listing}{Listing}
\usepackage[hang,flushmargin]{footmisc}

\title{GFlow: Recovering 4D World from Monocular Video}
\author{
    Shizun Wang, \,\, Xingyi Yang, \,\, Qiuhong Shen, \,\, Zhenxiang Jiang, \,\, Xinchao Wang\thanks{Corresponding author.}
}
\affiliations{
    National University of Singapore \\
    shizun.wang@u.nus.edu, xinchao@nus.edu.sg

}

\usepackage{bibentry}

\begin{document}


\twocolumn[{%
\renewcommand\twocolumn[1][]{#1}%
\maketitle
\vspace{-3em}
\begin{center}
    \includegraphics[width=\textwidth]{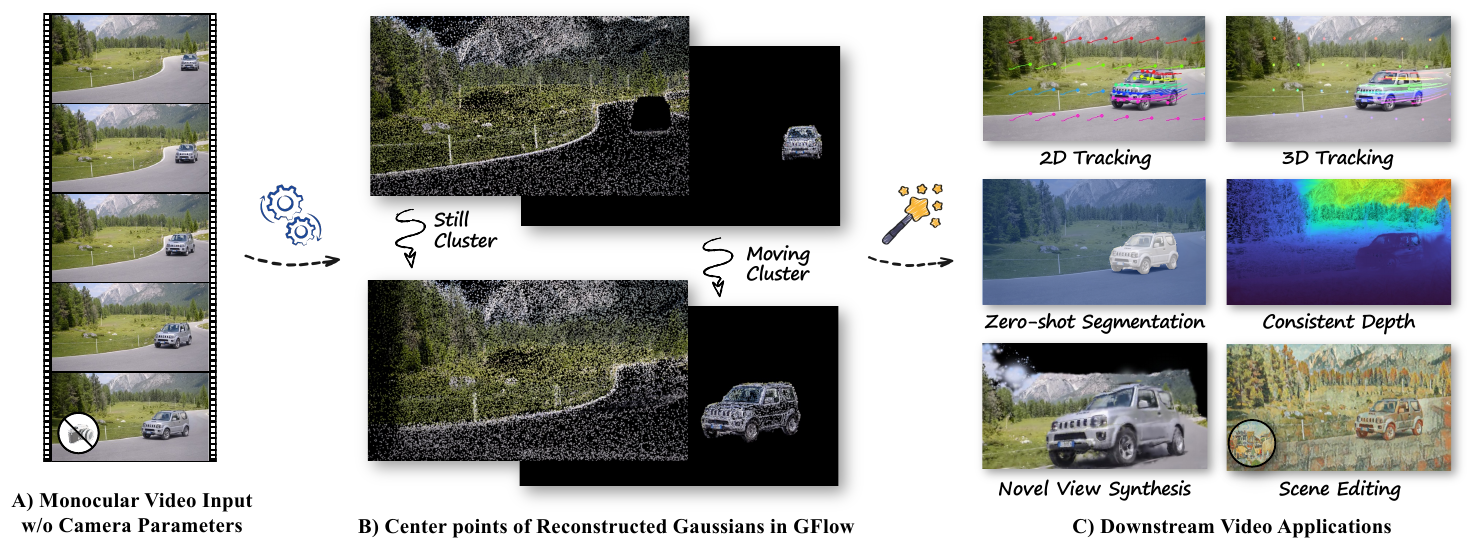}
    \captionof{figure}{A) Given a monocular video in the wild, B) our proposed \textbf{GFlow} can reconstruct the underlying 4D world, \textit{i.e.} the dynamic scene represented by 3D Gaussian splatting \cite{kerbl20233d} and associated camera poses. Within GFlow, the Gaussians are split into still and moving clusters and and are further densified. C) GFlow facilitates a range of applications, including tracking objects in 2D and 3D, segmenting video objects, synthesizing new views, estimating consistent depth and video editing. We encourage readers to visit the anonymous website for more video illustrations.}
    \label{fig:teaser}
\end{center}
}]

\begin{abstract}

\let\thefootnote\relax\footnotetext{\hspace{1em} \textsuperscript{\rm *}Corresponding Author}
\let\thefootnote\relax\footnotetext{Copyright \copyright \space 2025, Association for the Advancement of Artificial Intelligence (www.aaai.org). All rights reserved.}
  
Recovering 4D world from monocular video is a crucial yet challenging task. 
Conventional methods usually rely on the assumptions of multi-view videos, known camera parameters, or static scenes.
In this paper, we relax all these constraints and tackle a highly ambitious but practical task: With only one monocular video without camera parameters, we aim to recover the dynamic 3D world alongside the camera poses.
To solve this, we introduce \textbf{GFlow}, a new framework that utilizes only 2D priors (depth and optical flow) to lift a video to a 4D scene, as a flow of 3D Gaussians through space and time. 
GFlow starts by segmenting the video into still and moving parts, then alternates between optimizing camera poses and the dynamics of the 3D Gaussian points.
This method ensures consistency among adjacent points and smooth transitions between frames.
Since dynamic scenes always continually introduce new visual content, we present prior-driven initialization and pixel-wise densification strategy for Gaussian points to integrate new content. 
By combining all those techniques, GFlow transcends the boundaries of 4D recovery from causal videos; it naturally enables tracking of points and segmentation of moving objects across frames.
Additionally, GFlow estimates the camera poses for each frame, enabling novel view synthesis by changing camera pose. This capability facilitates extensive scene-level or object-level editing, highlighting GFlow's versatility and effectiveness.
\end{abstract}

%
\begin{links}
    \link{Website}{https://littlepure2333.github.io/GFlow}
\end{links}


\section{Introduction}

The quest for accurate reconstruction of 4D scene from video inputs stands at the forefront of contemporary research in computer vision and graphics. This endeavor is crucial for the advancement of virtual and augmented reality, video analysis, and multimedia applications. The main challenge lies in capturing the transient essence of dynamic scenes and the often absent camera pose information. 

Traditional approaches are typically split between two types: the one relies on pre-calibrated camera parameters or multi-view video inputs to reconstruct dynamic scenes \cite{wu20234d, luiten2023dynamic, sun20243dgstream, bansal20204d, cao2023hexplane, fridovich2023k, li2022neural, lin2023high, lin2021deep, pumarola2021d}, while the other estimates camera poses from static scenes using multi-view stereo techniques \cite{bian2023nope, fu2023colmap, wang2023dust3r, lin2021barf, wang2021nerf, xia2022sinerf, schonberger2016pixelwise, schonberger2016structure, tian2023mononerf, charatan2023pixelsplat}. This division highlights a missing piece in this field: the challenge of reconstructing dynamic scenes using only a single monocular video without any camera parameters.

Addressing this challenge is particularly difficult due to the inherently \emph{ill-posed} nature. From a single monocular video, multiple reconstructions might visually appear correct when projected onto the camera view. However, many of these reconstructions fail to conform to the physical laws of the real world. Although NeRF-based \cite{cao2023hexplane, fridovich2023k, shao2023tensor4d, liu2023robust} methods attempt to solve this problem, they often yield poor results. This failure is primarily due to their implicit representation, which makes it challenging to accurately enforce physical constraints in the reconstruction.

Recent developments in 3D Gaussian Splatting (3DGS) \cite{kerbl20233d} and its extensions \cite{wu20234d, luiten2023dynamic, yang2023deformable, yang2023real} into dynamic scenes have emerged as promising alternatives. 
These techniques have shown promise in handling the complexities associated with the dynamic nature of real-world scenes, as well as the intricacies of camera movement and positioning.Yet, they still operate under the assumption of known camera poses \cite{schonberger2016pixelwise, schonberger2016structure}.

To transcend these limitations and fully leverage the capabilities of 2D foundation models for dynamic scene reconstruction, we offer a novel insight:

\textbf{\textit{Given 2D factors such as RGB, depth and optical flow from one video, we have enough clues to model the 4D (3D+t) world behind the video.}}

Leveraging this insight, we introduce \textbf{GFlow}, a novel framework that leverages 3D Gaussian Splatting \cite{kerbl20233d} to reconstruct the video. It conceptualizes the video content as a dynamic flow of Gaussian points moving through space and time. We simultaneously optimize the flow and the camera poses together, to ensure that the projected video adheres to those 2D factors.

The key to GFlow lies in the alternating optimization of camera poses and dynamic 3D Gaussians. While directly estimating camera poses in dynamic scenes is considered highly challenging, we make it feasible by separating the scene into static and dynamic parts. For the static parts, we optimize the camera poses using reprojection error. In dynamic regions, Gaussian points are first reprojected using the optimized camera poses, then refined based on RGB, depth, and flow priors. This dual optimization ensures that each video frame is rendered accurately, capturing the dynamic nature of the original scene.

Apart from the optimization strategy, we propose two methods to effectively integrate new Gaussian points into the scene and accelerate convergence. The first method, \emph{ prior-driven initialization}, sets up initial Gaussian points in plausible 3D geometric positions, based on RGB and depth priors. The second method, \emph{pixel-wise densification}, involves increasing the number of Gaussian points in regions with large pixel errors. Together, these strategies contribute to maintaining high fidelity in cross-frame rendering, also ensuring that transitions and movements between frames are smooth.

Beyond dynamic 3D scene recovery, GFlow can also serve as a powerful tool for video processing. It can track points across frames in 3D world coordinates without prior training and segment objects by propagating a given initial mask. Since it employs explicit representation, GFlow can render captivating new views of video scenes by easily changing camera poses and editing objects or entire scenes as desired, showcasing its versatility and power.

To conclude, our contributions are: \textbf{1)} A novel framework that recovers 4D scenes and associated camera poses from a monocular video. \textbf{2)} An alternating optimization process that ensures high fidelity and temporally smooth dynamics in 4D scenes. \textbf{3)} Two new strategies for initializing and densifying Gaussian points in dynamic scenes. \textbf{4)} Enables new video processing capabilities, including tracking, segmentation, novel view rendering, and editing.

\section{Related works}

\subsubsection{3D Renderable Representations} 
Static 3D scenes can be recovered as renderable representations from posed multi-view images through differentiable rendering, enabling novel view synthesis. Such 3D renderable representations can be categorized into implicit and explicit representations. Early works in 3D scene reconstruction primarily adopted implicit neural representations~\cite{deepstereo}. The most influential of these, Neural Radiance Fields (NeRFs)~\cite{mildenhall2021nerf}, introduced importance sampling with volumetric ray-marching but relied on a deep multi-layer perceptron, significantly hindering rendering speed. Although follow-up works~\cite{instant-ngp, tensoRF} adopted hash grids or structured tensors with smaller MLPs to represent density and appearance, their rendering speed is still constrained by the need to query substantial samples for single ray marching.

In contrast, the explicit category is dominated by differentiable point-based rendering techniques~\cite{pointsurface, kerbl20233d}. This approach eliminates the need to query samples from deep networks, instead directly fetching attributes from points, which enables a significant speedup compared to implicit neural-based methods. Recently, 3D Gaussian Splatting (3DGS)~\cite{kerbl20233d} extends points to 3D Gaussians with opacity and spherical harmonics, and introduces tile-based rasterization to achieve real-time rendering speeds.
Here, we choose 3DGS as our base representation, as its fast rendering speed and the explicit nature make the reconstructed scene flexible enough for content creation and editing.

\subsubsection{4D Reconstruction}
4D reconstruction from video, also known as dynamic 3D scene reconstruction. Many prior works extended NeRFs to handle dynamic scenes~\cite{park2021nerfies, park2021hypernerf, pumarola2021d, li2023dynibar}, typically using grids, triplanes, voxels~\cite{cao2023hexplane, fridovich2023k, shao2023tensor4d, liu2023robust}, or learning deformable fields to map a canonical template \cite{ouyang2023codef, kasten2021layered}. But the reconstruction quality is relatively low due to its implicit essence.
Recent developments in 3DGS~\cite{kerbl20233d} have set new records in reconstruction quality and rendering speed. Extensions of 3DGS~\cite{wu20234d, luiten2023dynamic, yang2023deformable, yang2023real} have begun exploring dynamic scene reconstruction. However, they still operate under the assumption of a known camera sequence~\cite{schonberger2016pixelwise, schonberger2016structure}.

While almost all previous methods either rely on known camera parameters or multi-view video inputs to reconstruct dynamic scenes \cite{sun20243dgstream, bansal20204d, li2022neural, lin2023high, lin2021deep}, or estimate camera poses from static scenes using multi-view stereo techniques \cite{bian2023nope, fu2023colmap, wang2023dust3r, lin2021barf, wang2021nerf, xia2022sinerf, tian2023mononerf, charatan2023pixelsplat}. 
The key difference between our GFlow and these approaches lies in our ability to recover dynamic scenes from a unposed monocular video. 
Additionally, some concurrent works~\cite{wang2024shape, stearns2024dynamic, lei2024mosca, liu2024modgs, Konghanyang} also attempt to solve this problem.

\section{Preliminaries}
\subsection{3D gaussian splatting}

Recently, 3D Gaussian Splatting (3DGS) \cite{kerbl20233d} exhibits strong performance and efficiency in 3D representation. 3DGS fits a scene as a set of Gaussians $\{G_i\}$ from multi-view images $\{V_{k}\}$ and paired camera poses $\{P_{k}\}$ in an optimization pipeline. Adaptive densification and pruning of Gaussians are applied in this iterative optimization to control the total number of Gaussians. Generally, each Gaussian is composed of its center coordinate $\mu \in \mathbb{R}^3$, 3D scale $s \in \mathbb{R}^3$, opacity $\alpha \in \mathbb{R}$, rotation quaternion $q \in \mathbb{R}^4$, and associated view-dependent color represented as spherical harmonics $c \in \mathbb{R}^{3(d + 1)^2}$, where $d$ is the degree of spherical harmonics. 

These parameters can be collectively denoted by $G$, with $G_i = \{\mu_i, s_i, \alpha_i, q_i, c_i\}$ denoting the parameters of the $i$-th Gaussian. The core of 3DGS is its tile-based differentiable rasterization pipeline to achieve real-time optimization and rendering. To render $\{G_i\}$ into a 2D image, each Gaussian is first projected into the camera coordinate frame given the camera pose $P_i$ to determine the depth of each Gaussian. Then colors, depth, or other attributes in pixel space are rendered in parallel by alpha composition with the depth order of adjacent 3D Gaussians. Specifically, in our formulation, we do not consider view-dependent color variations for simplicity, thus the degree of spherical harmonics is set as $d = 0$, i.e., only the RGB color $c \in \mathbb{R}^3$.

\subsection{Camera model}
\label{sec:camera}
To project the 3D point coordinates $\mu \in \mathbb{R}^3$ into the camera view, we use the pinhole camera model. The camera intrinsics is $K \in \mathbb{R}^{3 \times 3}$ and the camera extrinsics which define the world-to-camera transformation is $E = [R|t] \in \mathbb{R}^{3 \times 4}$. The camera-view 2D coordinates $x \in \mathbb{R}^2$ are calculated as $d h(x) = K E h(\mu)$, where $d \in \mathbb{R}$ is the depth, and $h(\cdot)$ represents the homogeneous coordinate mapping.

\section{Methodology}

\paragraph{Problem Definition}

We aim to address a highly challenging and ill-posed problem, which is commonly encountered in real-world scenarios though:
Given a sequence of monocular video frames without known camera parameters, the objective is to model the dynamic 3D world and the associated camera poses to represent the video.

\begin{figure*}
  \centering
  \includegraphics[width=\linewidth]{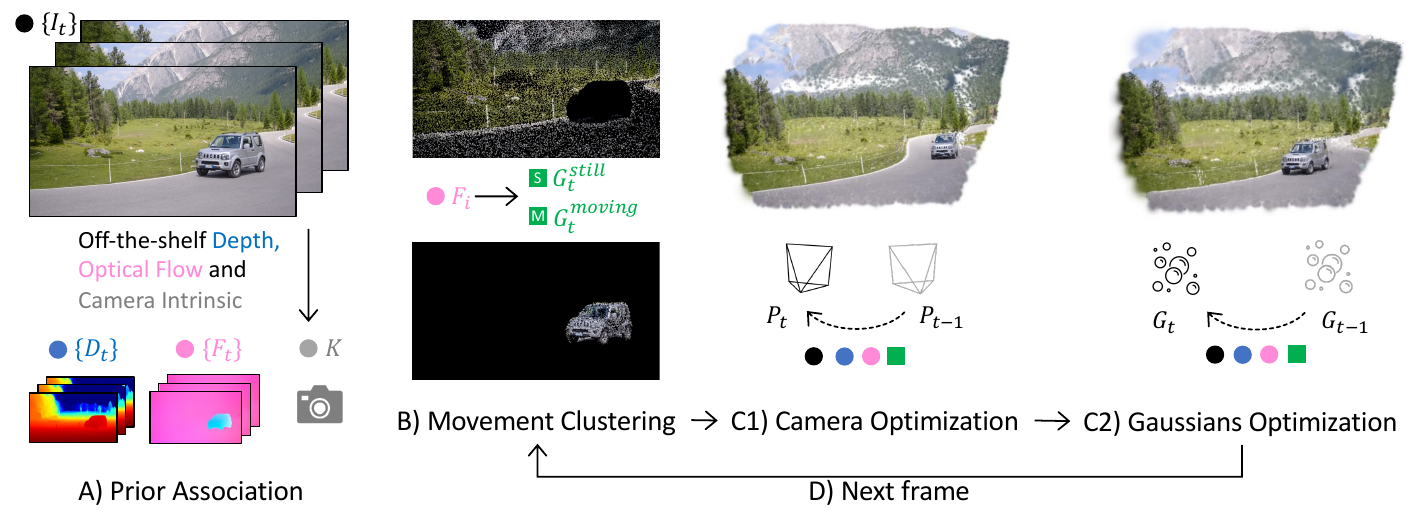}
  \caption{\textbf{Overview of GFlow.} \textbf{A)} Given a monocular video input consisting of image sequence $\{I_t\}$, the associated depth $\{D_t\}$, optical flow $\{F_t\}$ and camera intrinsic $K$ are obtained using off-the-shelf prior. \textbf{B)} For each frame , GFLow first clustering the scene into still part $\{G_t^{s}\}$ and moving part $\{G_t^{m}\}$. Then optimization process in GFlow consists of two steps: \textbf{C1)} Only the camera pose $P_t$ is optimized by aligning the appearance, depth and optical flow within the still cluster. \textbf{C2)} Under the optimized camera pose $P_t^*$, the Gaussian points $\{G_t\}$ are optimized and densified based on appearance, depth, optical flow and the two scene clusters. \textbf{D)} The same procedure of steps B, C1, and C2 loops for the next frame. The colorful marks under the dashed line represent the variables involved in the optimization.}
  \label{fig:overview}
\end{figure*}

\paragraph{Overview}

To address this problem, we propose \textbf{GFlow}, a framework that represents videos through a flow of 3D Gaussians, as shown in Figure~\ref{fig:overview}. We first preprocess the videos to derive several priors using advanced foundation models. The priors include depth \cite{leroy2024grounding}, optical flow \cite{xu2023unifying}, and camera intrinsics \cite{leroy2024grounding}, which we believe are the minimum necessary. These priors contribute to good initialization and regularization in the GFlow optimization process. Two novel strategies are devised to effectively deal with the Gaussian points initialization and densification in the dynamic scenes (Sec. \ref{sec:allocation}). At the essence of proposed method, GFlow alternately optimizes the camera pose and Gaussian points for each frame in sequential order to reconstruct the 4D world, assisted by movement clustering of Gaussian points (Sec. \ref{sec:alternating}).

\subsection{Gaussian Points Allocation}
\label{sec:allocation}

This section introduces new strategies for initializing and densifying Gaussian points according to the video content.

\subsubsection{Prior-driven Initialization of Gaussians}
\label{sec:ini}
The original 3D Gaussian Splatting \cite{kerbl20233d} initializes Gaussian points using point clouds derived from Structure-from-Motion (SfM) \cite{schonberger2016structure, schonberger2016pixelwise}, which are only viable for static scenes with dense views. However, our task involves dynamic scenes that change both spatially and temporally, making SfM infeasible.

To address this, we developed a new method called \textbf{prior-driven initialization} for single frames. This method fully utilizes the \emph{texture} information and \emph{depth} estimation obtained from the image to initialize the Gaussian points.

Intuitively, image areas with more edges usually indicate more complex textures, so more Gaussian points should be initialized in these areas. 
Given an image $I \in \mathbb{R}^{H \times W}$, we extract its texture map $T \in \mathbb{R}^{H \times W}$ the Sobel operator~\cite{kanopoulos1988design}, an edge detection operator. We then normalize this texture map to create a probability map $P \in \mathbb{R}^{H \times W}$, from which we sample $N$ points to obtain their 2D coordinates $\{x_i\}_{i=1}^N$. 

To obtain their position in the 3D space, we use depth $D$ estimated from off-the-shelf model \cite{leroy2024grounding}, as it can offer strong geometric information. The depth $\{d_i\}_{i=1}^N$ of sampled points can be retrieved from depth map $D$ using  2D coordinates. 
The 3D center coordinate $\{\mu_i\}_{i=1}^N$ of Gaussian points is initialized by unprojecting depth $\{d_i\}_{i=1}^N$ and camera-view 2D coordinates $\{x_i\}_{i=1}^N$, according to the pinhole model. 
The scale $\{s_i\}_{i=1}^N$ and color $\{c_i\}_{i=1}^N$ of the Gaussian points are initialized based on the probability values and pixel colors retrieved from the image, respectively.

\subsubsection{Pixel-wise Densification of Gaussians}
\label{sec:densification}
3D Gaussian Splatting~\cite{kerbl20233d}, designed for static scenes, uses gradient thresholding to densify Gaussian points; points exceeding a gradient threshold are cloned or split based on their scales. However, this method struggles in dynamic scenes, particularly when camera movements reveal new scene areas where no Gaussian points exist.

To address this, we introduce a new \textbf{pixel-wise densification} strategy that leverages image content, specifically targeting areas yet to be fully reconstructed.  Our approach utilizes a pixel-wise photometric error map $E_{pho} \in \mathbb{R}^{H \times W}$ and a mask $M_e \in \mathbb{R}^{H \times W}$ as the basis for densification. This masked error map is then normalized into a probability map $P_e \in \mathbb{R}^{H \times W}$. To densify new Gaussian points, the same initialization method described in prior-driven initialization is adopted, with the exception of sampling from $P_e \odot M_e $. The number of new Gaussian points introduced is proportionate to the mask ratio, ensuring controlled expansion of the point set.

There are two scenarios for densification: 1) Before Gaussian optimization, 
the mask $M_e$ only marks new content, which is detected via a forward-backward consistency check using bidirectional flow from advanced optical flow estimators \cite{xu2022gmflow, xu2023unifying}.
And we set the $P_e$ as uniform probabililty map, to fill new content emerged in a new frame. 
2) During Gaussian optimization, $P_e$ is unchanged, and the mask $M_e$ is identified by thresholding the error map $E_{pho}$, ensuring densification occurs at details-lacking area.

\subsection{Alternating Gaussian-Camera Optimization}
\label{sec:alternating}
Once the first frame has been initialized and optimized, for subsequent frames, we adopt a alternating optimization strategy for the camera poses $\{P_i\}$ and the Gaussians $\{G_i\}$.

\setcounter{footnote}{0} 

\subsubsection{Movement Clustering of Gaussian Points}
\label{sec:cluster}
In constructing dynamic scenes that include both camera and object movements, treating these scenes as static can lead to inaccuracies and loss of crucial temporal information. To better manage this, we propose a method to cluster the scene into still and moving parts, which will be incorporated in the optimization process.

We calculate the epipolar error map based on the optical flow estimated by UniMatch \cite{xu2023unifying, xu2022gmflow}. The moving mask $M_t$ at time $t$ is identified by thresholding the epipolar error map. When Gaussian points are initialized or densified, those within $M_t$ are considered as moving points $\{G_i^{m}\}_t \subseteq \{G_i\}_t$, while others are considered as still points $\{G_i^{s}\}_t \subseteq \{G_i\}_t$, whose center coordinate $\{\mu_i^s\}$ will stop updating.
This simple yet effective movement clustering method enables GFlow to model and track both rigid and non-rigid movement, whether it occurs on objects or other elements like water.

\subsubsection{Camera Optimization}
\label{sec:opt-camera}
The camera intrinsic $K$ is estimated by MASt3R \cite{leroy2024grounding}. Between two consecutive frames, only camera extrinsic $E$ is optimizable, all Gaussian points are frozen. The camera extrinsic $E = [R|t]$ consists of a rotation $R \in \mathbf{SO}(3)$ and a translation $t \in \mathbb{R}^3$.

\begin{figure*}[ht]
  \centering
  \includegraphics[width=\linewidth]{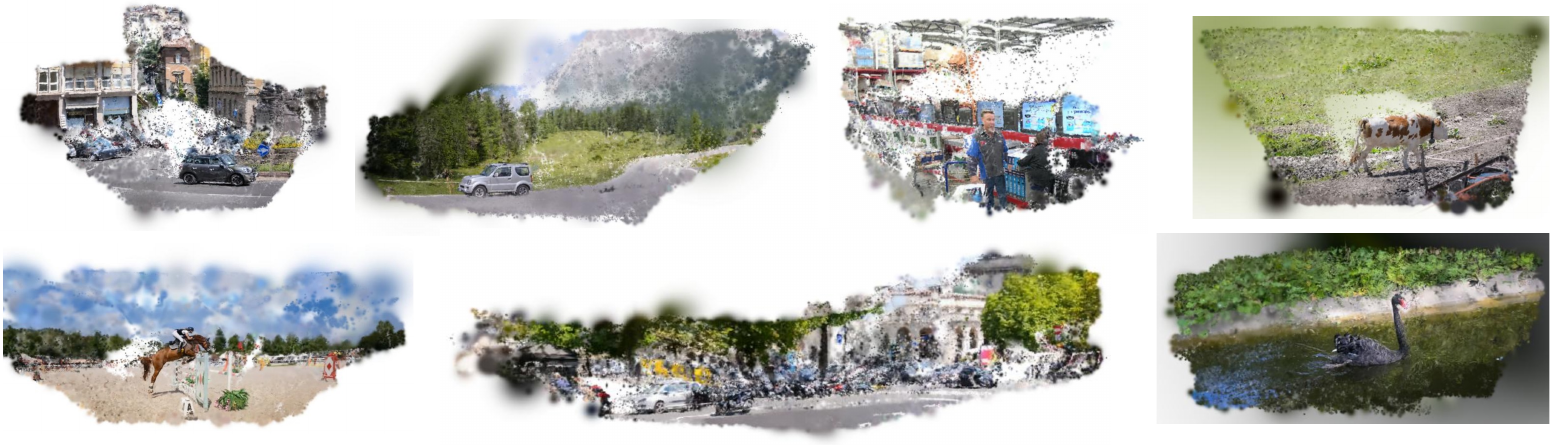}
  \caption{Our GFlow can explicitly model the dynamic 3D scene in the video. Here we show some rendered examples of videos from DAVIS \cite{perazzi2016benchmark, pont20172017} dataset in the 3D world space.}
  \label{fig:3d}
\end{figure*}

For a given frame at time $t$, we optimize the camera extrinsic $E_t$ by minimizing the errors in its photometric appearance, depth and optical flow. 

\begin{equation}
    E_t^* = \mathop{\arg\min}\limits_{E_t} \quad (\lambda_p \mathcal{L}_{pho} + \lambda_d \mathcal{L}_{dep} + \lambda_f \mathcal{L}_{flo}),
\end{equation}

where $\lambda_p$, $\lambda_d$ and $\lambda_f$ are weighting factors.
During this optimization, since the moving part is not contribute to camera pose estimation, we will mask out the moving area according to current and previous moving mask $M_t, M_{t-1}$.

Here, $\mathcal{R}(\cdot)$ denotes the 3D Gaussian splatting rendering process for desired output.
The \textit{photometric loss} combines MSE and SSIM~\cite{wang2004image} loss between the rendered image $\hat{I}_t = \mathcal{R}(\{G_i\}_t)$ and the actual frame image $I_t$. 

\begin{equation}
    \mathcal{L}_{pho} = \mathcal{L}_{mse} (\hat{I}_t, I_t) + \mathcal{L}_{ssim} (\hat{I}_t, I_t)
\end{equation}
\label{equ:photo}

The \textit{depth loss} is calculated using the L1 loss between the rendered depth $\hat{D}_t = \mathcal{R}(\{G_i\}_t)$ and prior depth $D_t$. 
To address discrepancies in scale and shift between the rendered and prior depths, we apply a scale and shift-invariant term on the loss, where $a$ and $b$ are optimizable.

\begin{equation}
\mathcal{L}_{dep} = | (a * \hat{D}_t + b) - D_t |
\end{equation}
\label{equ:dep}

The \textit{optical flow loss} is calculated using the mean squared error (MSE) loss between the movements of Gaussian points in camera view $\hat{F}_t$ and the prior optical flow $F_t$, to ensure the temporal smoothness of the Gaussian points' trajectories. 

\begin{equation}
    \mathcal{L}_{flo} = \mathcal{L}_{mse} (\hat{F}_t, F_t) 
\end{equation}
\label{equ:flow}

\subsubsection{Gaussians Optimization}
\label{sec:opt-gaussian}

Once the optimized camera extrinsics $E_t^*$ is obtained, we first conduct \textbf{pre-optimization gaussians relocation} for those moving Gaussian points $\{G_i^{m}\}_t $. 
Initially, we retrieve the 2D coordinates of moving Gaussian points from the previous frame $\{x(G_{i,t-1}^m)\}$. Using these coordinates, we calculate their movement based on the previous frame's optical flow map $\{F_{t-1}(x(G_{i,t-1}^m))\}$ and update their current position: $x(G_{i,t}^m)=x(G_{i,t-1}^m)+F_{t-1}(x(G_{i,t-1}^m))$. With the updated coordinates, we then extract the depth from the current frame's depth map $\{D_t(x(G_{i,t}^m))\}$, and project these points from the camera view to world coordinates using $E_t^*$. This step ensures that the moving Gaussian points are accurately positioned and serves as a good initilization for subsequent optimization.

Then, the total optimization objectives of Gaussian points contains photometric loss, depth loss, optical flow loss, and additional isotropic loss.

\begin{equation}
    \{G_i^*\}_t = \mathop{\arg\min}\limits_{\{G_i\}_t} \left(\lambda_p \mathcal{L}_{pho} + \lambda_d \mathcal{L}_{dep} + \lambda_f \mathcal{L}_{flo} + \lambda_i \mathcal{L}_{iso} \right)
\end{equation}

Where $\lambda_i$ is a weight factor. The \textit{isotropic loss} is calculated as the mean of the standard deviation of the Gaussian points' 3D scales $\{s_i\}$. 
In this monocular setting with sparse views, the Gaussians tend to elongate along the view ray direction due to the lack of sufficient multi-view constraints. Applying isotropic loss will encourage the Gaussians to be isotropic, helping to reduce needle-like artifacts.

\begin{equation}
    \mathcal{L}_{iso} = \frac{1}{N} \sum_{i=1}^{N} \text{std}(s_i)
\end{equation}
\label{equ:iso}

The photometric and isotropic loss is applied to all Gaussian points, whereas the depth and optical flow losses focus specifically on the moving cluster $\{G_i^{m}\}_t$.
This approach ensures tailored optimization that balances the dynamics and stability of Gaussian points in the scene.

\begin{figure*}[t]
  \centering
  \includegraphics[width=\linewidth]{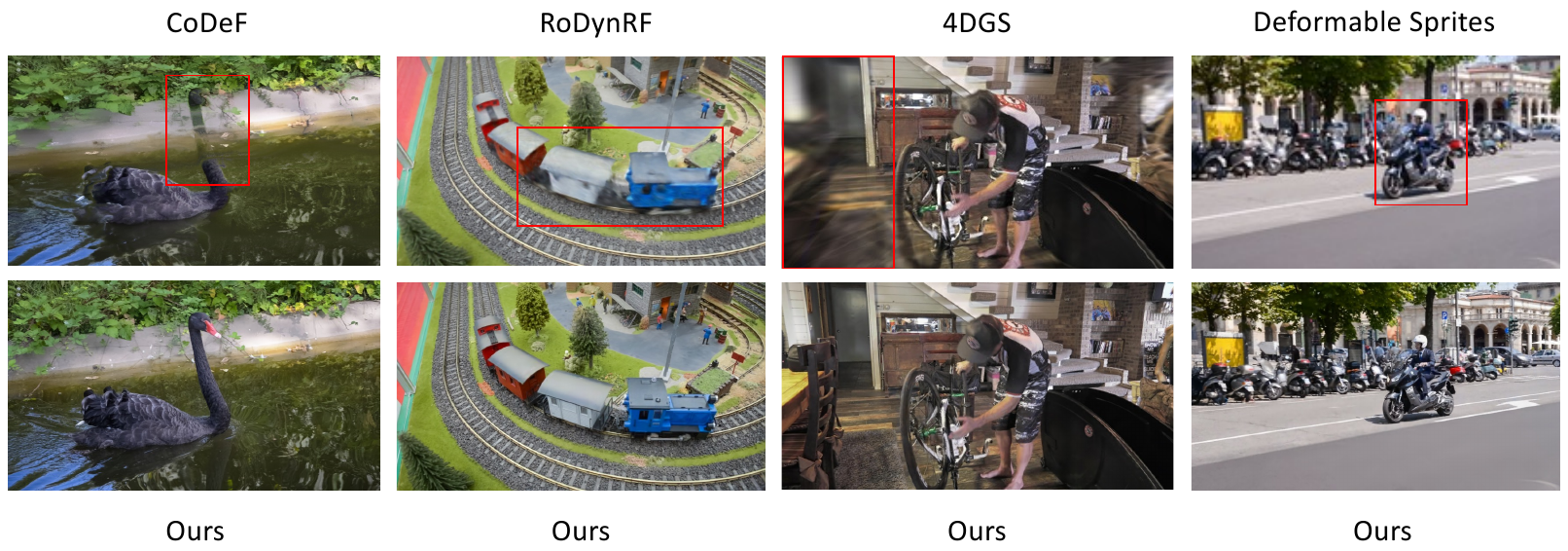}
  \caption{Visual comparison of reconstruction quality on the DAVIS \cite{perazzi2016benchmark, pont20172017} dataset: CoDef \cite{ouyang2023codef}, RoDynRF \cite{liu2023robust}, 4DGS \cite{yang2023gs4d}, and Deformable Sprites \cite{ye2022sprites} and Ours. }
  \label{fig:rec}
\end{figure*}

\subsection{Overall pipeline}
To conclude, the overall pipeline can be summarized as follows:
Given an image sequence $\{I_t\}_{t=0}^T$ of monocular video input, we first utilizes off-the-shelf algorithms \cite{xu2023unifying,xu2022gmflow, leroy2024grounding} to derive the corresponding depth $\{D_t\}_{t=0}^T$, optical flow $\{F_t\}_{t=0}^T$ and camera intrinsic $K$. 
The initialization of the Gaussians is performed using the prior-driven initialization.
Then for each frame $I_t$ at time $t$, GFlow first divides the Gaussian points $\{G_i\}_t$ into still cluster $\{G_i^{s}\}_t$ and moving cluster $\{G_i^{m}\}_t$ according to the optical flow.
The optimization process then proceeds in two steps. In the first step, only the camera extrinsics $E_t$ is optimized. This is achieved by aligning the Gaussian points within the still part with the appearance $I_t$, depth $D_t$ and optical flow $F_t$. Following that, under the optimized camera extrinsics $E_t^*$, the Gaussian points $G_t$ are further refined using constraints from the RGB $I_t$, depth $D_t$, optical flow $F_t$, and isotropic loss $\mathcal{L}_{iso}$. Additionally, the Gaussian points are densified using our proposed pixel-wise strategy to incorporate newly visible scene content. After optimizing the current frame, the procedure — movement clustering, camera optimization, and Gaussian point optimization — is repeated for subsequent frames. It is worth noting that in practice, GFlow is highly efficient, taking only 10 to 20 minutes to optimize a video — significantly faster than other methods that typically require more than an hour.

\section{Experiments}

\paragraph{Dataset and Metrics} 
We conduct experiments on a challenging video dataset, \textbf{DAVIS} \cite{perazzi2016benchmark, pont20172017} dataset, which contains real-world videos of about $30\sim100$ frames with various scenarios and motion dynamics. We report the reconstruction quality results on 30 DAVIS 2017 evaluation videos.
For quantitative evaluation of reconstruction quality, we report standard PSNR, SSIM and LPIPS~\cite{lpips} metrics. 

\paragraph{Implementation details}
All image sequences are resized to the shortest side is 480 pixels. The initial number of Gaussian points is set to 50,000. 
The camera intrinsics $K$ are estimated by MASt3R \cite{leroy2024grounding}.
The learning rate for Gaussian optimization is set to $4e\text{-}3$, and for camera optimization, it is set to $1e\text{-}3$. 
The Adam optimizer is used with 500 iterations for Gaussian optimization in the first frame, 150 iterations for camera optimization, and 300 iterations for Gaussian optimization in subsequent frames.
The gradient of color is set to zero, enforcing Gaussian points to move rather than lazily changing color.
We balance the loss term by setting $\lambda_p=1$, $\lambda_d=0.1$, $\lambda_f=0.01$, and $\lambda_i=50$. 
Densification is conducted at the 150 and 300 steps in the first frame optimization. 
For subsequent frames, the densification occurs at the first step with a new content mask applied, and also occurs at the 100 and 200 steps with error-thresholding mask applied. The error threshold in densification is set to 0.01.
All experiments are conducted on a single NVIDIA RTX A5000 GPU.
Note that the dynamics within each video could be distinct, so for better reconstruction, the hyperparameters could be tuned in practice.

\subsection{Evaluation of Reconstruction Quality}
\paragraph{Quantitative Results}
Reconstructing the 4D world, particularly with camera and content movement, is an extremely challenging task. 
We choose several methods closest to tackle this task as our baseline. 
Deformable Sprites \cite{ye2022sprites} decomposes the video into layers of persistent motion groups, which are then composed to reconstruct the video.
RoDynRF \cite{liu2023robust} uses neural voxel radiance fields to model the dynamic scene and camera poses simultaneously.
CoDeF \cite{ouyang2023codef} employs implicit content deformation fields to learn a canonical template for modeling monocular video, but it lacks physical interpretability, such as the ability to estimate camera poses.
4DGS \cite{yang2023gs4d} models the space and time dimensions for dynamic scenes by formulating unbiased 4D Gaussian primitives, though it requires camera poses as input. We use the camera poses estimated by MASt3R as its input.
As shown in Table \ref{tab:rec}, our GFlow demonstrates significant advantages in reconstruction quality. This improvement stems from its explicit representation and tailored optimization process design, which can adapt Gaussian points over time while maintaining visual content coherence.

\begin{table}
  \centering
  \begin{tabular}{llll}
    \toprule
    \multirow{2}[2]{*}{Method} & \multicolumn{3}{c}{DAVIS}   \\
    \cmidrule(lr){2-4} 
                               & PSNR$\uparrow$     & SSIM$\uparrow$    & LPIPS$\downarrow$   \\
    \midrule
    Deformable Sprites    &   22.83       &   0.6983      &    0.3014    \\
    RoDynRF    &   24.79       &   0.7230      &    0.3940    \\
    CoDeF    &   24.89       &   0.7703      &    0.2932    \\
    4DGS    &   24.60       &   0.7315      &    0.3710    \\
    \midrule
    \textbf{GFlow (Ours)}     &   \textbf{29.74}       &    \textbf{0.9205}     &   \textbf{0.1237}        \\
    GFlow*       &    29.21      &    0.9162     &    0.1320   \\
    w/o $\mathcal{L}_{dep}$       &    29.30      &    0.9086     &    0.1444   \\
    w/o $\mathcal{L}_{iso}$       &    26.32      &    0.8664     &    0.2003   \\
    \bottomrule
  \end{tabular}
  \caption{Reconstruction quality results on DAVIS\cite{perazzi2016benchmark, pont20172017} dataset. Average PSNR, SSIM and LPIPS scores on all videos are reported.}
  \label{tab:rec}
\end{table}

\paragraph{Qualitative Results}

The visual comparison in Figure~\ref{fig:rec} shows that CoDeF struggles to reconstruct videos with significant movement due to its reliance on representing a video as a canonical template. RoDynRF has difficulty reconstructing high-quality moving objects. Even with camera pose inputs, 4DGS falls short in reconstructing the entire frame image. Additionally, Deformable Sprites can only reconstruct videos at a very low resolution. In contrast, our GFlow can faithfully reconstruct both static and moving content in high quality. The visual illustration in Figure~\ref{fig:3d} demonstrates the dynamic 3D scene recovered from monocular videos, showcasing the effectiveness of our approach.

\subsection{Ablation study}

\subsubsection{Effect of isotropic loss}
Since the monocular video input only provides sparse and underconstrained views, traditional 3DGS, which relies on dense multi-view constraints, struggles to achieve good results. The sparse views will result in needle-like artifacts along the camera view ray direction. As shown in Table~\ref{tab:rec}, the isotropic loss helps to overcome these drawbacks and improves the reconstruction quality.

\subsubsection{Effect of depth loss}
Depth loss is used for ensuring a consistent and reasonable 3D geometry structure. Table~\ref{tab:rec} shows the reconstruction quality will decrease without depth loss.

\subsubsection{Effect of optimizing camera pose}
When preprocessing the monocular video using MASt3R \cite{leroy2024grounding}, it can also estimate camera poses. The results labeled as `GFlow*' in Table~\ref{tab:rec} show the effect of directly using the camera poses estimated by MASt3R instead of optimizing them. A decrease in quality is observed, demonstrating the necessity of optimizing camera poses. Additionally, GFlow is capable of optimizing camera poses even when most areas in the video are in motion, where MASt3R fails.

\subsection{Downstream video applications}

Various downstream applications can be extended from our GFlow framework, as it is an explicit representation. We encourage readers to view the videos in the website for more intuitionistic illustration.

\subsubsection{Point tracking}
\begin{figure}[t]
  \centering
  \includegraphics[width=\linewidth]{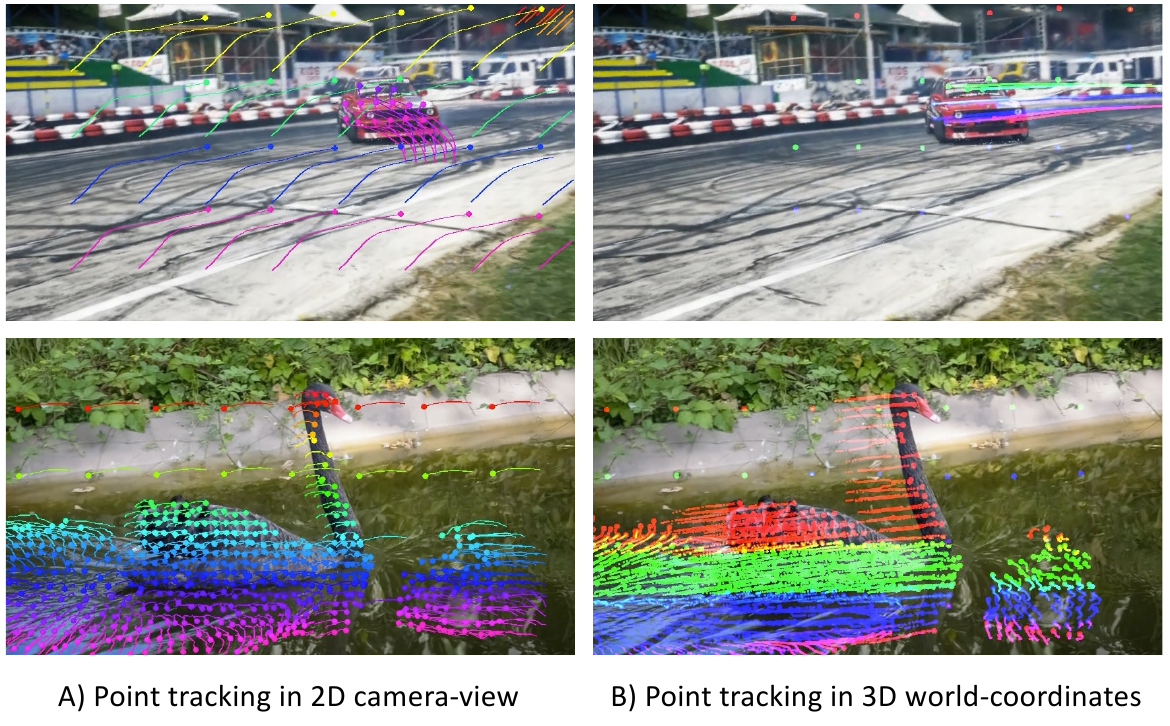}
  \caption{Point tracking visualization on DAVIS dataset. 
  A) tracking in the 2D camera-view which contains joint motion of camera and content movement. B) tracking in the 3D world-coordinates which only present content movement.}
  \label{fig:track}
\end{figure}

Due to the nature of GFlow, all Gaussian points can serve as query tracking points, enabling tracking in both 2D and 3D space. The tracking trajectories are illustrated in Figure~\ref{fig:track}.
In conventional \textbf{2D tracking}, tracking occurs in the camera view, which includes the combined motion of both the camera and the content. In contrast, the Gaussian points in GFlow reside in 3D world coordinates, representing only content movement. As a result, some \textbf{3D tracking} trajectories tend to remain in their original locations, as shown in Figure~\ref{fig:track}B), because they belong to the static background. These results demonstrate that GFlow can achieve accurate tracking even on water ripples and remains reliable for fast-moving objects and scenes.

\subsubsection{Video Object Segmentation}
Since GFlow drives the Gaussian points to follow the movement of the visual content, given an initial mask, all Gaussian points within this mask can propagate to subsequent frames. This propagation forms a new mask as a concave hull \cite{park2012new} around these points. Notably, this capability is a by-product of GFlow, achieved without extra intended training.

\subsubsection{Video Editing}

Since explicit representation can be easily edited:
\textbf{Camera-level manipulation}: 
Changing the camera extrinsics can render novel views of dynamic scenes. When combined with camera intrinsics, it can create visual effects like dolly zoom.
\textbf{Object-level editing}: 
With the cluster labels of moving Gaussian points, we can add, remove, resize, or stylize these points, allowing for precise object-level editing.
\textbf{Scene-level editing}:
Editing can also be applied to the entire scene, enabling the application of visual effects globally, as illustrated in Figure~\ref{fig:teaser}.

\section{Conclusion}
We have presented ``GFlow", a novel framework designed to address the challenging task of reconstructing the 4D world from casual monocular videos. Through Gaussian points allocation and alternating camera-Gaussian optimization, GFlow enables the recovery of dynamic scenes alongside camera poses across frames. Further capabilities such as tracking, segmentation, editing, and novel view rendering, highlighting GFlow's potential to revolutionize video understanding and manipulation.

\section*{Acknowledgments}
This project is supported by the National Research Foundation, Singapore, under its Medium Sized Center for Advanced Robotics Technology Innovation, and the Singapore Ministry of Education Academic Research Fund Tier~1 (WBS: A-0009440-01-00).

\bibliography{aaai25}

\appendix

\section{More Implementation Details}

\subsubsection{Preprocessing}
We employ UniMatch \cite{xu2023unifying, xu2022gmflow} for optical flow estimation. Specifically, we utilize the scale-2 model, which incorporates an additional 6 local regression refinement steps and is trained on a mixture of public datasets, making it well-suited for in-the-wild scenarios. For depth estimation and camera intrinsics, we adopt MASt3R \cite{leroy2024grounding, wang2023dust3r}, performing the estimation at a subsample-2 scale with a shared intrinsic across all frames.

\subsubsection{Initialization}
Once we obtain the probability map $P$, we normalize all non-zero values. Based on this probability map , the 3D center coordinates of Gaussian points are initialized by unprojecting the depth along camera-view 2D coordinates. The scale of Gaussian points is initialized as the odds of probability, then scaled by a factor of corresponding depth to ensure suitability for the screen size. The color of Gaussian points is initialized using the color retrieved from the image corresponding to the camera-view 2D coordinates. 
The opacity is initialized as 0.99, and the rotation is randomly initialized.

Once the probability map \( P \) is obtained, we normalize all non-zero values. Using this probability map, the 3D center coordinates of the Gaussian points are initialized by unprojecting the depth along the camera-view 2D coordinates. The scale of the Gaussian points is initialized as the odds of the probability and then scaled by a factor of the corresponding depth to ensure suitability for the screen size. The color of the Gaussian points is initialized using the color retrieved from the image at the corresponding camera-view 2D coordinates. The opacity is initialized as 0.99, and the rotation is randomly initialized.

\subsubsection{Densification}
During densification, the number of new Gaussian points is determined by \( N_{\text{den}} = R_m \times N_{\text{ini}} \), where \( R_m \) represents the mask ratio (the ratio of the masked area to the total area of the frame), and \( N_{\text{ini}} \) denotes the initial number of Gaussian points, which is set to 50,000 in our experiments. 
There are two types of densification masks:  
1) New content mask: Used before Gaussian optimization, this mask is detected through a forward-backward consistency check based on bidirectional optical flow \cite{xu2022gmflow, xu2023unifying}.  
2) Under-reconstructed mask: Used during Gaussian optimization, this mask is obtained by thresholding the photometric error map \( E_{\text{pho}} \) with a threshold of 0.01.

\subsubsection{Movement Clustering}
Since the optical flow constraint operates in 2D space, multiple Gaussian points may exist along the 2D camera view ray in 3D space, especially when occlusion occurs. Therefore, after movement clustering, we freeze all center coordinates ${\mu_i}$ of static points ${G_i^s}_t$ to prevent them from being displaced by optical flow. The movement mask is identified by thresholding the epipolar error map, with the threshold set to $0.01$.

\begin{table}
  \centering
  \begin{tabular}{lccc}
    \toprule
    Method   & $\textit{ATE}\downarrow$  &  $\textit{RPE}_t\downarrow$ & $\textit{RPE}_r\downarrow$  \\
    \midrule
    R-CVD     &   0.360       &    0.154     &   3.443       \\
    DROID-SLAM         &   0.175       &   0.084      &   1.912      \\
    COLMAP      &   \multicolumn{3}{c}{\textcolor{gray}{Fails 5 out of 14 sequences}}    \\
    \midrule
    NeRF - -     &   0.433       &   0.220      &   3.088      \\
    BARF         &   0.447       &   0.203      &   6.353      \\
    RoDynRF      &   \textbf{0.089}       &   \underline{0.073}      &   \underline{1.313}      \\
    \midrule
    \textbf{GFlow}   &  \underline{0.124}      &   \textbf{0.039}      &    \textbf{0.599} \\
    \bottomrule
  \end{tabular}
  \caption{  Camera pose estimation results on the MPI Sintel dataset \cite{sintel}, reporting both Absolute Trajectory Error (ATE) and Relative Pose Error (RPE). The best results are highlighted in bold, while the second-best results are underlined. The methods in the top section can only estimate camera poses, do not reconstruct scene view images.}
  \label{tab:cam}
\end{table}

\begin{figure}[t]
  \centering
  \includegraphics[width=\linewidth]{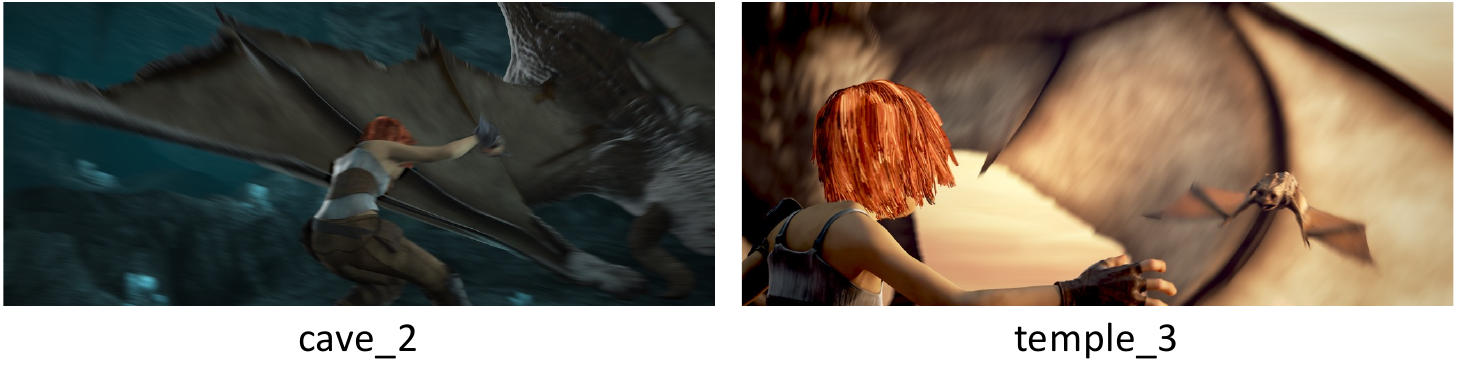}
  \caption{Some challenging cases from the MPI Sintel \cite{sintel} dataset include heavily occluded static backgrounds and significant motion blur, both of which complicate the camera optimization process.}
  \label{fig:sintel}
\end{figure}

\subsubsection{Camera Optimization}
When optimizing the camera pose, only the static part serves as a reference, and the moving part does not contribute or even be harmful to pose estimation. Therefore, we need to exclude the moving part, \( \overline{M} = M_t \cup M'_{t-1} \), which is the union of the current moving mask \( M_t \) and the previous moving mask \( M'_{t-1} \) in the new view. The mask \( M'_{t-1} \) is determined as follows: first, identify the previous moving Gaussian points \( \{G_i^m\} \) within the previous moving mask \( M_{t-1} \) from the previous frame. Then, splat these points \( \{G_i^m\} \) using the optimized camera pose \( E^* \) to get the moving part image \( \hat{I}^m \). Finally, threshold the grayscale image of the moving part image \( grey(\hat{I}^m) \) with a threshold of 0 to obtain the mask \( M'_{t-1} \).

\begin{figure*}[t]
  \centering
  \includegraphics[width=0.98\linewidth]{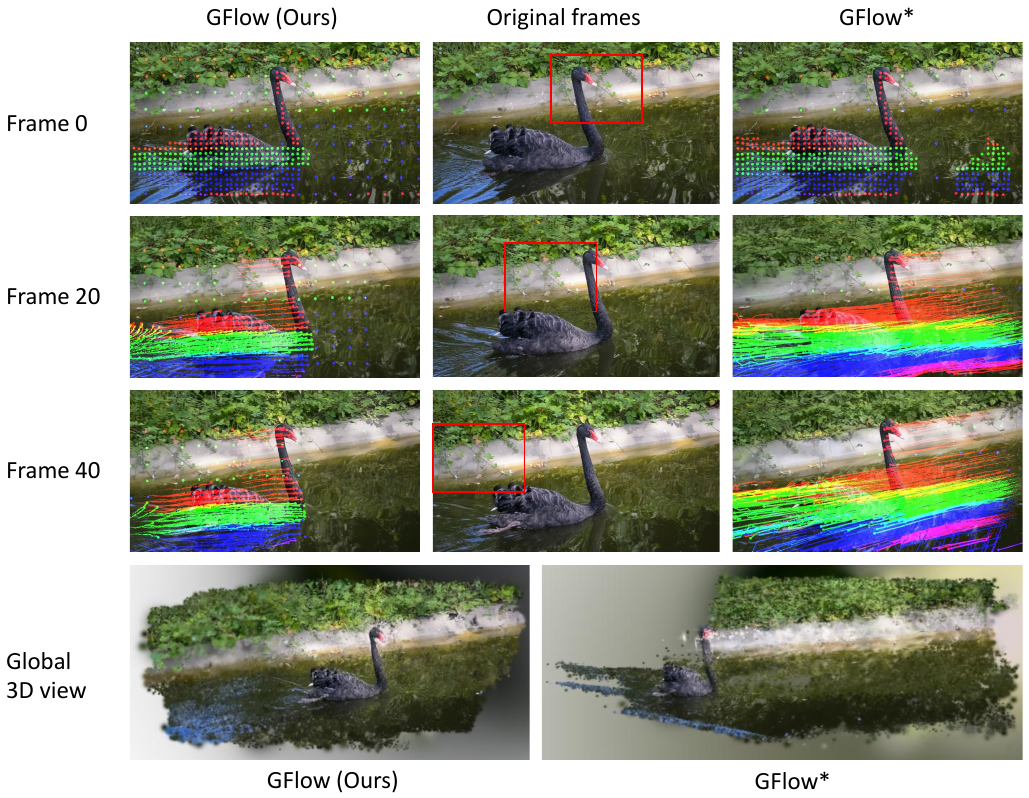}
  \caption{
  3D point tracking visualization example (blackswan) on the DAVIS \cite{perazzi2016benchmark, pont20172017} dataset. 
  The colorful dots and trajectories indicates the movement in 3D world coordinates.
  The red box helps the reader identify the reference plants in the background. The camera is moving to the right, and the black swan is also moving to the right in the video.}
  \label{fig:track_cam}
\end{figure*}

\subsubsection{Optimization Choices}
It is worth noting that, although all experiments follow the same hyperparameter settings, the results can be further improved by optimizing these settings for each specific case. This is reasonable because the dynamics and content of videos vary significantly (e.g., Figure~\ref{fig:sintel}).

For example, for a video with large camera motion, we can increase the learning rate or extend optimization iterations in the camera optimization process. For videos with clear and simple rigid motions, we can directly use the estimated camera poses \cite{leroy2024grounding} instead of optimizing them to shorten the overall processing time. 

\section{More Experiments}

\subsection{Evaluation of Camera Pose Estimation}
\subsubsection{Dataset and Metrics}
MPI Sintel \cite{sintel} dataset provides high-quality, synthetic sequences with complex motion, realistic lighting, and challenging visual effects like motion blur and depth of field. Following prior works~\cite{liu2023robust}, we evaluate 14 sequences with ground-truth camera poses provided.
As for camera pose accuracy, we report standard visual odometry metrics \cite{sturm2012benchmark, zhang2018tutorial}, including the Absolute Trajectory Error (ATE) and Relative Pose Error (RPE) of rotation and translation as in \cite{bian2023nope, lin2021barf}.

\subsubsection{Results}

Our method can reconstruct the 4D world along with the corresponding camera poses. The MPI Sintel dataset presents highly challenging dynamics which makes camera pose estimation more difficult, such as large occlusion by moving objects and severe motion blur. To alleviate these challenges, we combine the strengths of both optimizing camera poses and using estimated camera poses \cite{leroy2024grounding} as initialization. So we combines the best results from both settings. 

In Table \ref{tab:cam}, we compare the camera pose estimation results with R-CVD \cite{kopf2021robust}, DROID-SLAM \cite{teed2021droid}, COLMAP \cite{schoenberger2016sfm}, NeRF-- \cite{wang2021nerf}, BARF \cite{lin2021barf}, and RoDynRF \cite{liu2023robust}. Our method, GFlow, achieves comparable or better results in camera pose estimation compared to previous methods, demonstrating its effectiveness.

\subsection{Ablation study}
\subsubsection{Effect of optimizing camera pose}

As described in the main text, we report the reconstruction results between `GFlow (ours)' (optimizing camera pose) and `GFlow*' (directly using the camera poses estimated by MASt3R). Although the numerical differences are not significant, the camera pose estimation and movement are incorrect in the 4D world. We show an example in Figure \ref{fig:track_cam}. In this video, a black swan is swimming to the right on a river, and the camera is also moving to the right. The tracking trajectories and global 3D view of `GFlow (ours)' show the correct moving direction and camera poses, while `GFlow*' fails. Due to the flowing water, most of the frames are moving, making it difficult for MASt3R to estimate the correct camera poses. Please refer to the supplementary videos for a clearer illustration.

\end{document}